\documentclass[12pt]{article}

\usepackage[T1]{fontenc}
\usepackage{textcomp}
\usepackage{sbc-template}
\usepackage{graphicx,url}
\usepackage[utf8]{inputenc}
\usepackage[english]{babel}
\usepackage[flushleft]{threeparttable}
\usepackage{subfigure}
\usepackage{xcolor}
\usepackage{svg}
\usepackage{adjustbox}
\usepackage{amssymb} 
\usepackage{graphicx}
\usepackage{subcaption}
\usepackage{multirow}
\usepackage{amsmath} 
\usepackage[font=bf]{caption}
\usepackage{booktabs}
\usepackage{caption}

\newcommand{\eric}[1]{\textcolor{black}{#1}}
\newcommand{\eniac}[1]{\textcolor{black}{#1}}
\newcommand{\eniaccorrecao}[1]{\textcolor{black}{#1}}

\title{An Initial Study of
Bird's-Eye View Generation for Autonomous Vehicles 
using Cross-View Transformers
}

\author{
Felipe Carlos dos Santos\inst{1}, 
Eric Aislan Antonelo\inst{1}, 
Gustavo Claudio Karl Couto\inst{1}
}
\address{
Federal University of Santa Catarina (UFSC), Florianópolis, Brazil \\
Department of Automation and Systems Engineering 
\email{felipe.carlos.santos@posgrad.ufsc.br}, 
\email{eric.antonelo@ufsc.br, gustavo.karl.couto@posgrad.ufsc.br}
}
\begin{document}
%
%
%
\maketitle              
\begin{abstract}
%
%
Bird’s‑Eye View (BEV) maps provide a structured, top‑down abstraction that is crucial for autonomous‑driving perception. In this work, we employ Cross-View Transformers (CVT) for learning to map camera images to three BEV's channels - road, lane markings, and planned trajectory - using a realistic simulator for urban driving.
Our study examines generalization to unseen towns, the effect of different camera layouts, and two loss formulations (focal and L1). 
Using training data from only a town, a four‑camera CVT trained with the L1 loss delivers the most robust test performance, evaluated in a new town. Overall, our results underscore CVT’s promise for mapping camera inputs to reasonably accurate BEV maps.

\end{abstract}
%

%
%
%
%
%
%
\section{Introduction}\label{sec:intro}


Bird’s-eye-view (BEV) representations have become a cornerstone of modern autonomous driving stacks, because they remove perspective distortions of ego-centric cameras and expose the geometry of the scene in a planner-friendly grid.  
Early convolutional pipelines such as \textit{Lift-Splat-Shoot}~\cite{philion2020lss} showed that dense BEV features learned directly from multi-camera images boost both map segmentation and motion forecasting, triggering a surge of BEV-centric perception research.

Rather than rely on hand-crafted inverse-perspective mapping or depth supervision, recent methods employ Transformers to learn the view transformation implicitly.  
\textit{Cross-View Transformers (CVT)}~\cite{park2023cvt} introduced camera-aware attention that projects multi-view features into a unified map at real-time speed.  
Subsequent architectures—\textit{BEVFormer} with spatio-temporal queries~\cite{bevformer2022}, and the highly efficient \textit{WidthFormer}~\cite{widthformer2024} - have steadily advanced accuracy, latency and robustness on large-scale datasets such as nuScenes.

Dedicated lane- and drivable-area models now exploit decomposed or bidirectional cross-attention—e.g. \textit{Efficient Lane Transformer}~\cite{elt2023} and \textit{BAEFormer}~\cite{baeformer2023}—to overcome the geometric errors of naïve \textit{Inverse Perspective Mapping (IPM)} and to capture long-range context.  

Semantic-level BEV generation is also pivotal for closed-loop control.  
Imitation-learning frameworks such as \textit{Learning by Cheating}~\cite{codevilla2019cheating} demonstrate that mid-level BEV inputs can substantially stabilize end-to-end driving in CARLA, but they consume oracle or high-fidelity maps; the impact of a \emph{predicted} BEV on downstream control is somewhat unexplored, with early attempts made in \cite{Couto2023}.
Works in BEV generation can be broadly divided into two categories based on agent communication and sensor modality. The first category includes single-agent approaches, such as CVT, \textit{SinBEV} \cite{2022cobevt}, \textit{Fiery} \cite{hu2021fiery}, and \textit{VPN}  \cite{vpn2020}. The second category comprises multi-agent techniques that leverage vehicle-to-vehicle (V2V) communication to share information among agents, as seen in \textit{CoBEVT} \cite{2022cobevt}, \textit{V2VNet} \cite{wang2020v2vnet}, and \textit{CoBEVFusion} \cite{2024cobevfusion}.
In terms of sensor modality, BEV generation methods can be classified into those using only RGB cameras—such as CVT, Fiery and VPN—and those that rely on LiDAR or multi-modal inputs that combine various sensors, such as \cite{2024cobevfusion} and \cite{wang2020v2vnet}.

\eniaccorrecao{A BEV route channel is used by an imitation learning agent in \cite{antonelo2024investigating} to enable route-directed navigation. In \cite{Couto2023}, BEV generation via a GAN is conditioned on the intended route, allowing a downstream controller to use this information for navigation in the city. Similarly, we include a sparse trajectory as an additional input channel, enabling the network to predict a BEV route channel, which can be used by navigation agents in future work. }

\eric{This paper proposes the use of CVT in a single-vehicle approach that is based on camera input only, with training data from only one town.
}
The exclusive use of RGB cameras is motivated by the potential for lower hardware costs. Moreover, single-agent methods are particularly relevant in transitional scenarios where autonomous vehicles are not yet widespread, thus contributing to the practical viability and scalability of autonomous driving technologies.

\subsection{Contributions.}
This paper fills these gaps by:
\begin{enumerate}
  \item Adapting Cross-View Transformers to scenarios using the CARLA simulator, where \eric{it needs to map three frontal cameras and an optional backward view to three} semantic channels (road, lanes and planned route). 
  \eric{The generation of the planned route channel is particular to the current work, which was not covered in the literature as far as the authors know;}
  \item And benchmarking CVT against a strong UNet baseline under 
  \eric{equivalent training protocols},
  analysing data efficiency, view-count scalability (\eric{how gracefully a model’s performance changes when you add or remove camera views}) and generalization to unseen towns.
\end{enumerate}

\vspace{-0.5em}


\section{Methods}
\label{sec:methods}


\subsection{Unet}
Unet is a pure convolutional neural network, first introduced in \cite{unet2015} for medical image segmentation. Unet has become one of the most widely used architectures. It consists of an encoder that reduces spatial resolution through successive convolutional and pooling layers, capturing high-level semantic features, followed by a decoder that recovers spatial details via transposed convolutions, enabling precise pixel-wise predictions. In addition, Unet is known for using skip connections, a technique that has substantially improved the training of very deep neural networks \cite{orhan2017skip} by mitigating the vanishing gradient problem.   

While this architecture is originally designed for single-image input, our approach extends its use by concatenating multiple camera views alongside a trajectory representation. Although the network lacks an explicit cross-view fusion mechanism, this input configuration enables the CNN to jointly process information from all views, allowing it to generate a coherent semantic segmentation in the BEV view.

\subsection{Cross-view transformer (CVT)}
\label{sec:cvt_exp}
Following the success of the Transformer architecture introduced by \cite{transformers2017} in natural language processing, \cite{vit2020} demonstrated how the attention mechanism could be effectively adapted to process visual data, laying the groundwork for Vision Transformers. Building upon these concepts, \cite{park2023cvt} introduced a method that assigns a positional encoding to each individual camera based on its intrinsic and extrinsic parameters, enabling the fusion of multi-camera inputs into a unified map-view representation.

This model processes a collection of \( n \) monocular views \( \{(I_k, K_k, R_k, t_k)\}_{k=1}^n \), which include the input image \( I_k \in \mathbb{R}^{H \times W \times 3} \), camera intrinsic parameters \( K_k \in \mathbb{R}^{3 \times 3} \), along with the extrinsic rotation matrix \( R_k \in \mathbb{R}^{3 \times 3} \) and translation vector \( t_k \in \mathbb{R}^{3} \), all referenced to the ego-vehicle's center.


\subsubsection{Cross-view attention}

Given a real-world coordinate \( \mathbf{x}^{(W)} \in \mathbb{R}^3\), the perspective transformation defines its corresponding image-plane coordinate \( \mathbf{x}^{(I)} \in \mathbb{R}^3 \):

\begin{equation}
\mathbf{x}^{(I)} \sim K_k R_k (\mathbf{x}^{(W)} - \mathbf{t}_k).
\label{eq:image_coord}
\end{equation}

Using the image coordinates from Equation~\ref{eq:image_coord}, it is possible to rephrase the geometric relationship between real world and image coordinates as a cosine similarity, suitable for use in an attention mechanism, as follows:

\begin{equation}
\text{sim}_k(\mathbf{x}^{(I)}, \mathbf{x}^{(W)}) = 
\frac{
(R_k^{-1} K_k^{-1} \mathbf{x}^{(I)}) \cdot (\mathbf{x}^{(W)} - \mathbf{t}_k)
}{
\| R_k^{-1} K_k^{-1} \mathbf{x}^{(I)} \| \cdot \| \mathbf{x}^{(W)} - \mathbf{t}_k \|
}.
\label{eq:cosine_sim}
\end{equation}

This similarity still depends on the exact world coordinate \( \mathbf{x}^{(W)} \). To address this, the geometric components is replaced with positional encoding capable of capturing both geometric and appearance features.

\subsubsection{Camera-aware positional encoding}

Starting from the unprojected coordinate in the image world \(
\mathbf{d}_{k,i} = R_k^{-1} K_k^{-1} \mathbf{x}^{(I)}_i
\)
for each image coordinate \( \mathbf{x}^{(I)}_i \).

The vector \( \mathbf{d}_{k,i} \) is encoded through a linear layer using a multilayer perceptron (MLP), shared across all \( k \) views, to obtain a \( D \)-dimensional positional embedding.

The CVT architecture combines this embedding with the image features \( \boldsymbol{\phi}_{k,i} \) in the keys of the cross-view attention mechanism. This enables the attention module to perceive both appearance and geometric cues to establish correspondences across multiple views.



\subsubsection{Overall architecture}
\begin{figure}[h!]
    \centering
    \includegraphics[width=0.95\textwidth]{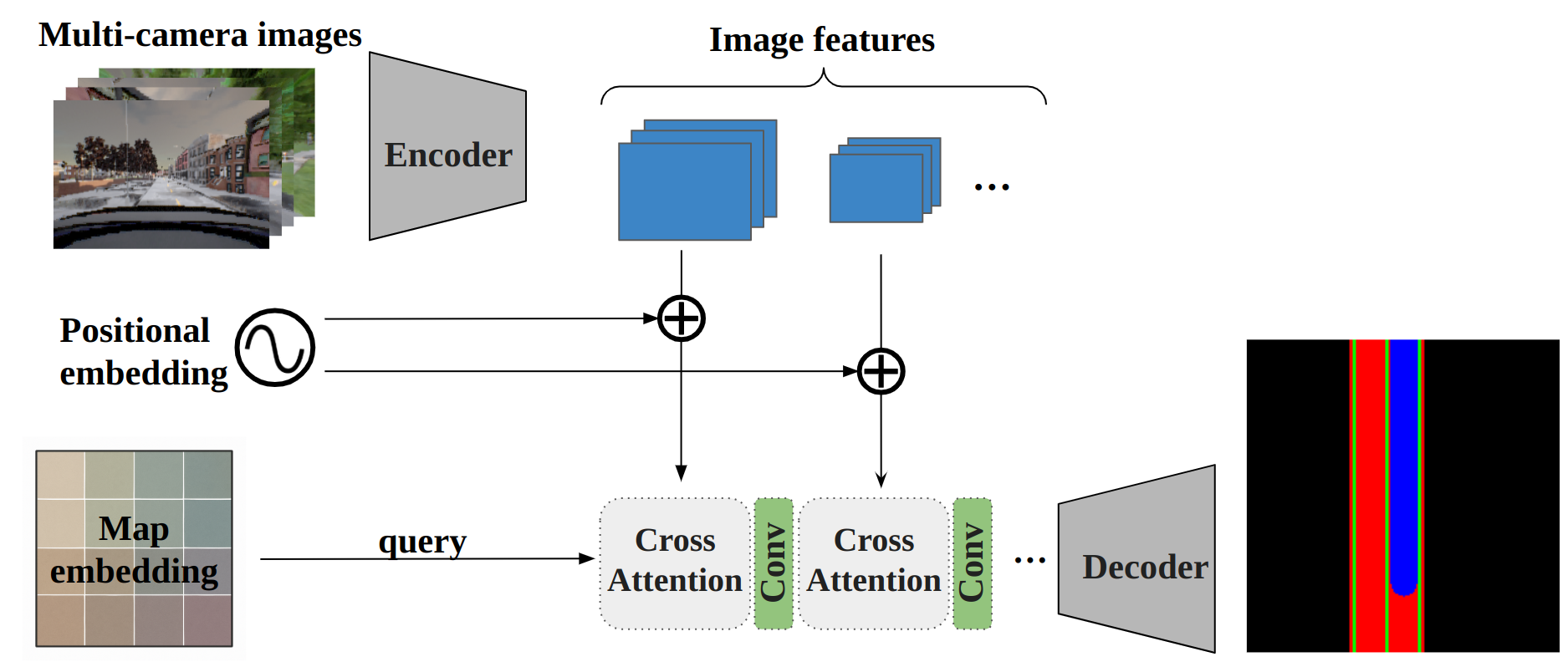} 
    \caption{\footnotesize \eniac{Architecture of the Cross-View Transformer (CVT). Multi-view images are processed through EfficientNet-B4 to extract multi-scale image features. 
    Camera-aware positional encodings and cross-view attention fuse geometric and visual information, enabling iterative refinement of the BEV representation in the latent space.}}
    \label{fig:cvt-diagram}
\end{figure}

The Figure~\ref{fig:cvt-diagram} shows the complete CVT architecture, each image \( I_i \) is passed through a feature extractor (EfficientNet-B4 \cite{tan2019efficientnet}) to produce a multi-resolution patch embedding
\(
\{\boldsymbol{\phi}^1_i, \boldsymbol{\phi}^2_i, \ldots, \boldsymbol{\phi}^R_i\},
\) where \( R \) is the number of resolutions considered, following \cite{park2023cvt} number of R used is 2. Then, the map-view representation is progressively refined and the process is repeated at higher resolutions. Finally, the decoder upsamples the representation to produce the final prediction.

\section{Experiments}
\label{sec:experiments}

In this section, we compare the Cross-View Transformer architecture with the Unet model, and evaluate the performance of two loss functions: L1 loss and Focal Loss \cite{lin2017focal}. Additionally, for the CVT, we compare configurations using four cameras and three cameras.

\subsection{Dataset Generation}
The environment and trajectories are sourced from the CARLA Leaderboard evaluation platform \cite{leaderboard_2020}. \eric{Specifically, from \textit{Town01} we build the training and validation sets, while from \textit{Town02} we create the test set. Each town provides ten different routes.}

\begin{figure}[ht]
    \centering
    \captionsetup{font=small, labelfont=bf}
    
    \begin{tabular}{@{}c@{\hspace{15mm}}c@{}}
        \includegraphics[width=0.45\linewidth]{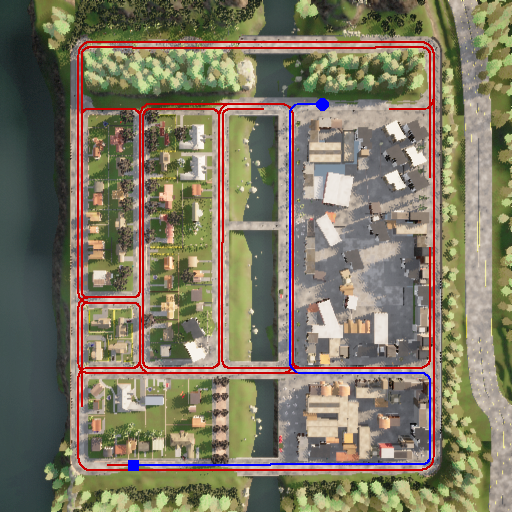} &
        \includegraphics[width=0.45\linewidth]{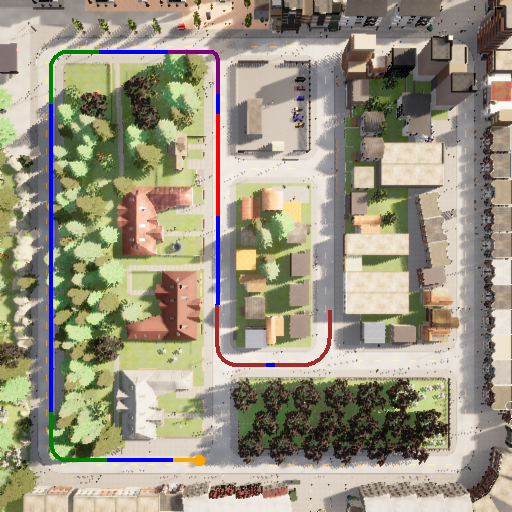} \\
        \scriptsize  (a) & 
        \scriptsize  (b) \\
    \end{tabular}

    \caption{\eniac{ (a) CARLA Leaderboard \textit{Town01}, showing training routes in red and the validation route in blue. The starting point is marked with a circle, and the endpoint is indicated with a square.
    (b) Test environment \textit{Town02}, with colored lines indicating different segments of \textit{route00}.
    }
    }
    \label{fig:test_envs}
\end{figure}

\begin{figure}[ht]
    \centering
    \captionsetup{font=small, labelfont=bf}
    
    \begin{tabular}{@{}c@{\hspace{1mm}}c@{\hspace{1mm}}c@{}}
        \includegraphics[width=0.25\linewidth]{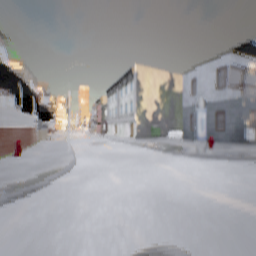} &
        \includegraphics[width=0.25\linewidth]{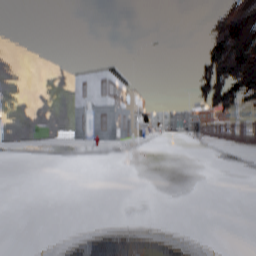} &
        \includegraphics[width=0.25\linewidth]{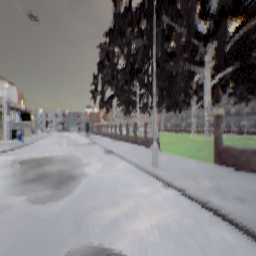} \\
        \scriptsize (a) & \scriptsize (b) & \scriptsize (c) \\
    \end{tabular}
    
    \vspace{0.5cm}
    
    \begin{tabular}{@{}c@{\hspace{15mm}}c@{}}
        \includegraphics[width=0.25\linewidth]{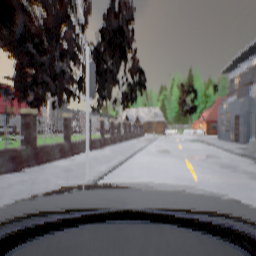} &
        \includegraphics[width=0.25\linewidth]{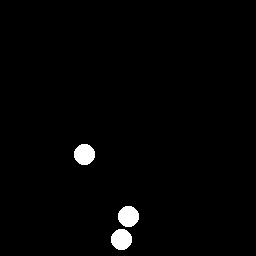} \\
        \scriptsize (d) & \scriptsize (e) \\
    \end{tabular}
    \caption{
        Example of model input captured in \textit{Town02}: 
        (a) Left camera, 
        (b) Central camera, 
        (c) Right camera, 
        (d) Rear camera, 
        (e) Sparse trajectory.
    }
    \label{fig:input_example}
\end{figure}

\eric{
The test \textit{Town02} is shown in Figure~\ref{fig:test_envs}(b), while
Figure~\ref{fig:input_example} shows the frames captured in that city at a particular instant by all the four cameras of the vehicle along with a sparse trajectory, totaling five input channels that feed the learning model. This input was later used to produce the inferences shown in Figure~\ref{fig:combined_1391}.
For the three-camera experiment, the rear camera is removed to analyze the model's behavior under limited information from the surroundings.}

Figure~\ref{fig:test_envs}(a) shows the \textit{Town01} environment, from which the nine training routes (red) and the validation route (blue) were obtained. On the other hand, Figure~\ref{fig:test_envs}(b) presents \textit{Town02}, used as the \eniac{test} environment, where the colored lines indicate segments of \textit{route00}.

    


\eniac{CVT requires the use of both intrinsic and extrinsic camera matrices, as discussed in Section~\ref{sec:cvt_exp},} but CARLA does not provide a straightforward way to retrieve them directly. Therefore, based on the camera configuration parameters, the intrinsic matrix used in this work is defined as follows. The focal lengths \( f_x \) and \( f_y \) are computed from the camera's field of view (FOV)—the angle through which the camera is capable of seeing the world—using the following equations. The image center coordinates \( c_x \) and \( c_y \) are approximated as \( c_x = \frac{\text{width}}{2} \) and \( c_y = \frac{\text{height}}{2} \):

\begin{equation}
f_x = \frac{\text{width}}{2 \cdot \tan\left(\frac{\text{FOV}_\text{rad}}{2}\right)}, \quad
f_y = \frac{\text{height}}{2 \cdot \tan\left(\frac{\text{FOV}_\text{rad}}{2}\right)}
\label{eq:focal_lengths}
\end{equation}

Given these values, the intrinsic matrix \( K \) from Section~\ref{sec:cvt_exp} is defined as:

\begin{equation}
K =
\begin{bmatrix}
f_x & 0 & c_x \\
0 & f_y & c_y \\
0 & 0 & 1
\end{bmatrix}
\label{eq:intrinsic_matrix}
\end{equation}

The extrinsic matrix is shown in Equation~\ref{eq:extrinsic_matrix}, where \( R \in \mathcal{R}^{3 \times 3} \) is the rotation matrix calculated from the \textit{pitch}, \textit{yaw}, and \textit{roll} parameters used to define the camera, and \( t \in \mathbb{R}^{3 \times 1} \)  is the position \( x, y, z \), which also defines the camera in the simulator. Both \( R \) and \( t \) are the same matrices discussed in Section~\ref{sec:cvt_exp}.

\begin{equation}
\mathbf{E} = \begin{bmatrix}
\mathbb{R} & \mathbb{t} \\
0 & 1
\end{bmatrix}
\label{eq:extrinsic_matrix}
\end{equation}

In total, the dataset comprises 233{,}315 images and 46{,}663 sample points, resulting in approximately 9.3~GB of uncompressed data. The distribution across the training, validation, and test sets is as follows:

\begin{itemize}
    \item \textbf{Training:} 96{,}435 images and 19{,}287 sample points, totaling approximately 3.8~GB.
    \item \textbf{Validation:} 10{,}000 images and 2{,}000 sample points, totaling approximately 0.3~GB.
    \item \textbf{Test:} 126{,}880 images and 25{,}376 sample points, totaling approximately 5.1~GB.
\end{itemize}


\subsection{Training}
\begin{figure}[h!]
    \centering
    \includegraphics[width=1\textwidth]{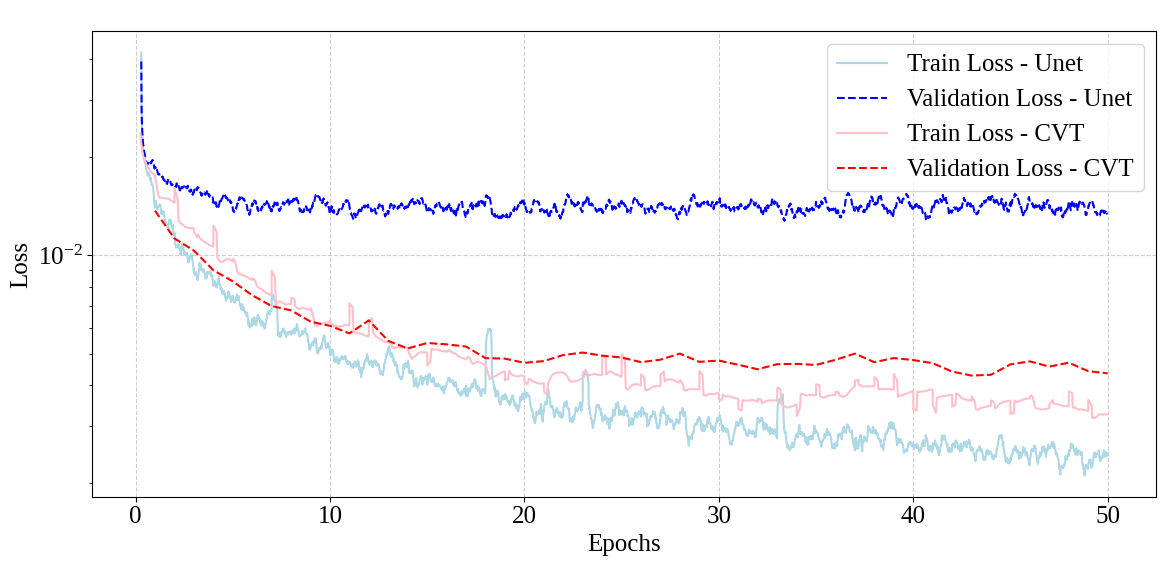} 
    \caption{
    \eric{Training and validation losses per epoch (\textit{Town01}) for both CVT and Unet using 4 cameras and focal loss.}}

    \label{fig:train_cvt_unet}
\end{figure}

Both 
\eric{Unet and CVT}
architectures were trained using 19{,}287 data points from \textit{routes 01} to \textit{09} in \textit{Town01}, while \textit{route00} was reserved for validation during training. All networks presented in this paper were trained for 50 epochs\footnote{The setting of 50 training epochs seamed reasonable enough to compare all the models, given the limited computational resources. Longer training periods with early stopping will be used in future works.} using the same learning rate configuration. 
\eniac{For the sake of simplicity, only the training losses for four cameras using focal loss are presented.}
Figure~\ref{fig:train_cvt_unet} shows the training metrics for CVT and UNet, both trained with focal loss and using four input cameras. Although the training loss for CVT is higher than that of UNet, the validation loss on \textit{route00} is 156.15\% higher for UNet. This result suggests that the transformer-based architecture has a stronger generalization capability, providing initial evidence of its superior performance.
\eric{performance on unseen input frames.}

In this work, six different models were trained: four based on the CVT architecture, using permutations of three (\eniac{no rear view}) and four input cameras combined with either focal loss or L1 loss; and two based on Unet, which serve as baselines, both using four cameras—one trained with focal loss and the other with L1 loss.

\subsection{Inference visualization}
\begin{figure}[ht]
    \centering

   \begin{minipage}{0.24\textwidth}
    \centering
    \begin{tabular}{@{}c@{}}
        \includegraphics[width=\linewidth]{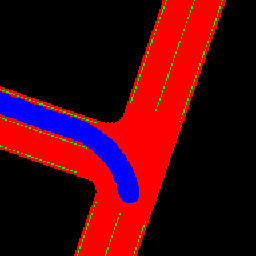} \\
        \textbf{Ground Truth}
    \end{tabular}
\end{minipage}
    \hfill
    \begin{minipage}{0.75\textwidth}
        \centering
        \begin{tabular}{@{}ccc@{}}
            \includegraphics[width=0.3\linewidth]{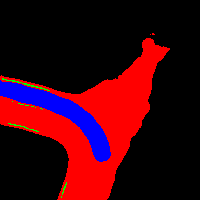} &
            \includegraphics[width=0.3\linewidth]{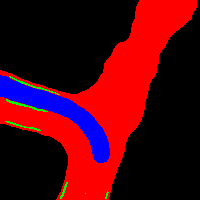} &
            \includegraphics[width=0.3\linewidth]{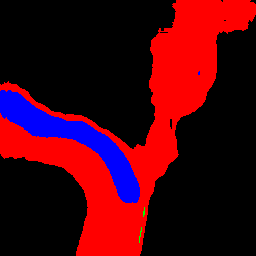} \\
             CVT-Focal 3 cams &  CVT-L1 3 cams &  CVT-Focal 4 cams
        \end{tabular}
    \end{minipage}

    \vspace{0.5cm}
    
    \begin{minipage}[t]{0.2\textwidth}
    \end{minipage}
    \hfill
    \begin{minipage}{0.75\textwidth}
        \centering
        \begin{tabular}{@{}ccc@{}}
            \includegraphics[width=0.3\linewidth]{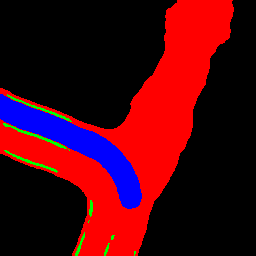} &
            \includegraphics[width=0.3\linewidth]{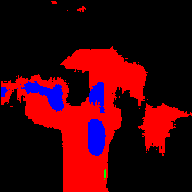} &
            \includegraphics[width=0.3\linewidth]{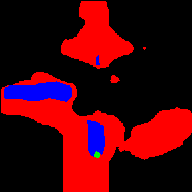} \\
             CVT-L1 4 cams &  UNet-Focal 4 cams &  UNet-L1 4 cams
        \end{tabular}
    \end{minipage}

    \caption{\eniac{Model inference results across all evaluated methods. The associated input images that served as input to these models are provided in Figure~\ref{fig:input_example}.}}
    \label{fig:combined_1391}
\end{figure}

Figure~\ref{fig:combined_1391} presents the inference results at a T-intersection in \textit{Town02} for each of the models.
\eric{Note that the models were trained only in \textit{Town01}, and that \textit{Town02} is used as an test town.}


\eric{From Figure~\ref{fig:combined_1391}, it can be observed that, as expected, the CVT architecture demonstrates superior BEV generation compared to Unet. However, the addition of the rear camera appears to introduce additional complexity into the model: without training it longer (all networks were trained for 50 epochs) neither with more data points, the performance of the models with 4 cameras decays compared to the its counterpart of 3 cameras. The longer training of these models, and with more data, is left as future work.
}

The combination of CVT with L1 loss shows promising results, particularly in the road channel, suggesting that this configuration may better capture the structural features of the environment. In the next subsection, we quantitatively analyze the model inferences in both \textit{Town01} and \textit{Town02}, as well as investigate how the models perform across different route segments.

\begin{figure}[ht]
    \centering
    
    \begin{minipage}{0.9\textwidth}
        \centering
        \textbf{Ground Truth} \\ 
        \vspace{0.2cm} 
        \begin{tabular}{@{}c|c|c|c@{}}
            \includegraphics[width=0.23\linewidth]{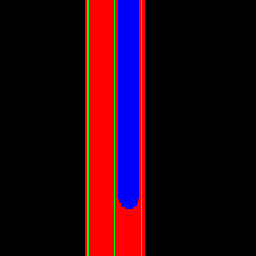} &
            \includegraphics[width=0.23\linewidth]{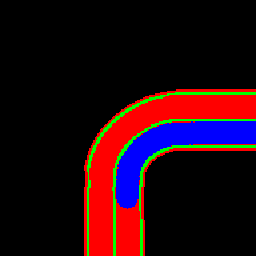} &
            \includegraphics[width=0.23\linewidth]{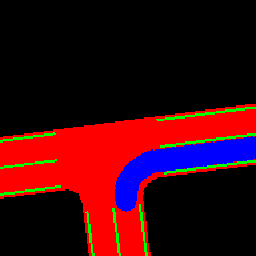} &
            \includegraphics[width=0.23\linewidth]{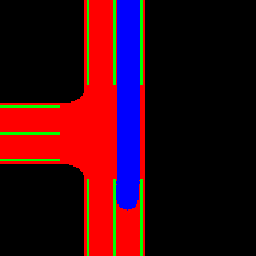}
        \end{tabular}
    \end{minipage}

    \vspace{0.5cm} 
    
    \begin{minipage}{0.9\textwidth}
        \centering
        \textbf{Inferences for CVT, L1 - 4 cams} \\ 
        \vspace{0.2cm} 
        \begin{tabular}{@{}c|c|c|c@{}}
            \includegraphics[width=0.23\linewidth]{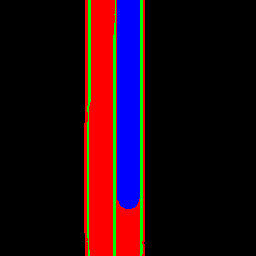} &
            \includegraphics[width=0.23\linewidth]{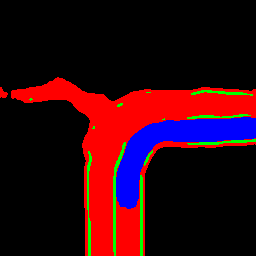} &
            \includegraphics[width=0.23\linewidth]{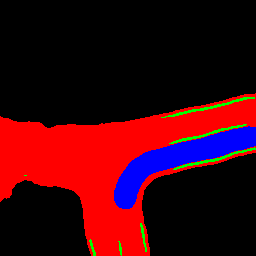} &
            \includegraphics[width=0.23\linewidth]{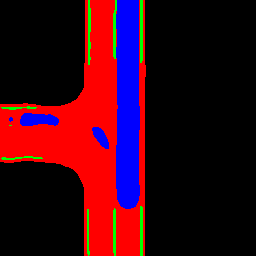} \\
             (a) &  (b) &  (c)  & (d)
        \end{tabular} 
    \end{minipage}

    \caption{\eniac{Model inferences for CVT 4 cameras trained using L1 loss, results across different locations in the test dataset - \textit{Town02}.}}
    \label{fig:l1_multiview}
\end{figure}


From Figure~\ref{fig:combined_1391} its possible to infer CVT outperforms UNet. The combination of CVT with L1 loss yields the best qualitative overall result. This figure also allows for a comparison between configurations with and without the rear camera. While the addition of the rear camera improves lane segmentation for the L1 loss configuration, this improvement is not observed with focal loss. 

Given the similar inferences observed in Figure~\ref{fig:combined_1391}, Table~\ref{tab:iou_best_models} presents the individual channel performances for these inferences. The table also shows that adding a rear camera in the focal loss configuration significantly improved the trajectory channel performance for this sample point.

\eniac{Figure~\ref{fig:l1_multiview} shows different trajectories for the CVT model with 4 cameras, trained using L1 loss. Figure~\ref{fig:l1_multiview}(a) displays a straight segment—the most common type—and the model performs very well in this case. Figures~\ref{fig:l1_multiview}(b), (c), and (d) present less frequent segments, where the model achieves relative success but struggles to segment the lane channel and blind spot regions.
}

\begin{table}[h!]
    \centering
    \caption{\centering mIoU per channel for the CVTs models from Figure ~\ref{fig:combined_1391} inferences}
    \label{tab:iou_best_models}
    \setlength{\tabcolsep}{12pt} 
    \begin{tabular}{l|c|c|c}
        \toprule
        \textbf{Model} & \textbf{Road} & \textbf{Trajectory} & \textbf{Lane} \\
        \midrule
        CVT-L1 4 cams     & 0.8188 & 0.6536 & \textbf{0.0355} \\
        CVT-L1 3 cams     & \textbf{0.8346} & 0.6890 & 0.0224 \\
        CVT-Focal 4 cams  & 0.7580 & \textbf{0.8051} & 0.0067 \\
        CVT-Focal 3 cams  & 0.6950 & 0.6709 & 0.0174 \\
        \bottomrule
    \end{tabular}
\end{table}


\subsection{Evaluation results}

To evaluate the trained models, two tables were generated, each comparing the mean IoU for every output channel across the different configurations. Table~\ref{tab:iou_tw1} presents the results on 2,000 data points from the validation route, \textit{route00}, in \textit{Town01}, while Table~\ref{tab:iou_tw2} shows the performance across all routes in \textit{Town02}, encompassing 25,376 data points. This comparison provides insights on how well the models generalize to environments and scenarios not seen during training.

\begin{table}[h!]
    \centering
    \caption{Mean IoU per channel for different models from \textit{Town01} - Validation route}
    \label{tab:iou_tw1}
     \setlength{\tabcolsep}{12pt} 
    \begin{tabular}{l|c|c|c}
        \toprule
         \textbf{Model} & \textbf{Road} & \textbf{Trajectory} & \textbf{Lane} \\
        \midrule
        CVT, Focal loss - 4 cams      & 0.9659 & 0.8650 & 0.4013 \\
        Unet, Focal loss - 4 cams     & 0.9072 & 0.7823 & \textbf{0.6972} \\
        CVT, L1 - 4 cams           & \textbf{0.9731} & 0.8563 & 0.4265 \\
        Unet, L1 - 4 cams               & 0.8863 & 0.7759 & 0.6591 \\
        CVT, Focal loss - 3 cams        & 0.9676 & \textbf{0.9028} & 0.4321 \\
        CVT, L1 - 3 cams           & 0.9593 & 0.8959 & 0.4317 \\
        
        \bottomrule
    \end{tabular}
    \
\end{table}
Table~\ref{tab:iou_tw1} shows that the CVT models demonstrate superior performance in representing both the road and trajectory channels when compared to the Unet baseline. On the other hand, the Unet model trained with focal loss achieves significantly better results in the lane channel, with an IoU around 60\% higher than the best-performing CVT model in that category.
%
\eric{For the road channel, the best model was the CVT with four cameras and L1 loss, but the three cameras model was slightly behind.}
However, for the trajectory channel, the addition of the rear camera seems to have negatively impacted performance, with the three-camera CVT variants outperforming their four-camera counterparts.
\begin{table}[h!]
    \centering
    \caption{\eniac{Per-channel mIoU scores for all evaluated models, tested on all \textit{Town02} ten routes.}}
    \label{tab:iou_tw2}
     \setlength{\tabcolsep}{12pt} 
    \begin{tabular}{l|c|c|c}
        \toprule
        \textbf{Model} & \textbf{Road} & \textbf{Trajectory} & \textbf{Lane}  \\
        \midrule
        CVT, Focal loss - 4 cams            & 0.9079 & 0.7528 & 0.2906 \\
       Unet, Focal loss - 4 cams      & 0.6978 & 0.5915 & \textbf{0.3959} \\
        CVT, L1 - 4 cams            & \textbf{0.9144} & 0.7808 & 0.3158 \\
        Unet, L1 4 - cams               & 0.6804 & 0.5642 & 0.3809 \\
        CVT, Focal loss - 3 cams     & 0.8981 & 0.7787 & 0.2970 \\ 
        CVT, L1 - 3 cams        & 0.8823 & \textbf{0.7935} & 0.3099 \\
        \bottomrule
    \end{tabular}
\end{table}

Table~\ref{tab:iou_tw2} presents the results for \textit{Town02}, where the trends observed in \textit{Town01} are largely maintained. 
\eric{The addition of the rear camera still slightly improves the performance for the road channel.}
For the trajectory channel, the top result was achieved by the CVT model with three cameras and L1 loss, closely followed by its four-camera counterpart. This suggests a slight advantage of L1 loss over focal loss in terms of generalization to unseen data. Notably, the same configuration also achieved the best CVT performance in the lane channel. 

\begin{figure}[h!]
    \centering
    \includegraphics[width=1\textwidth]{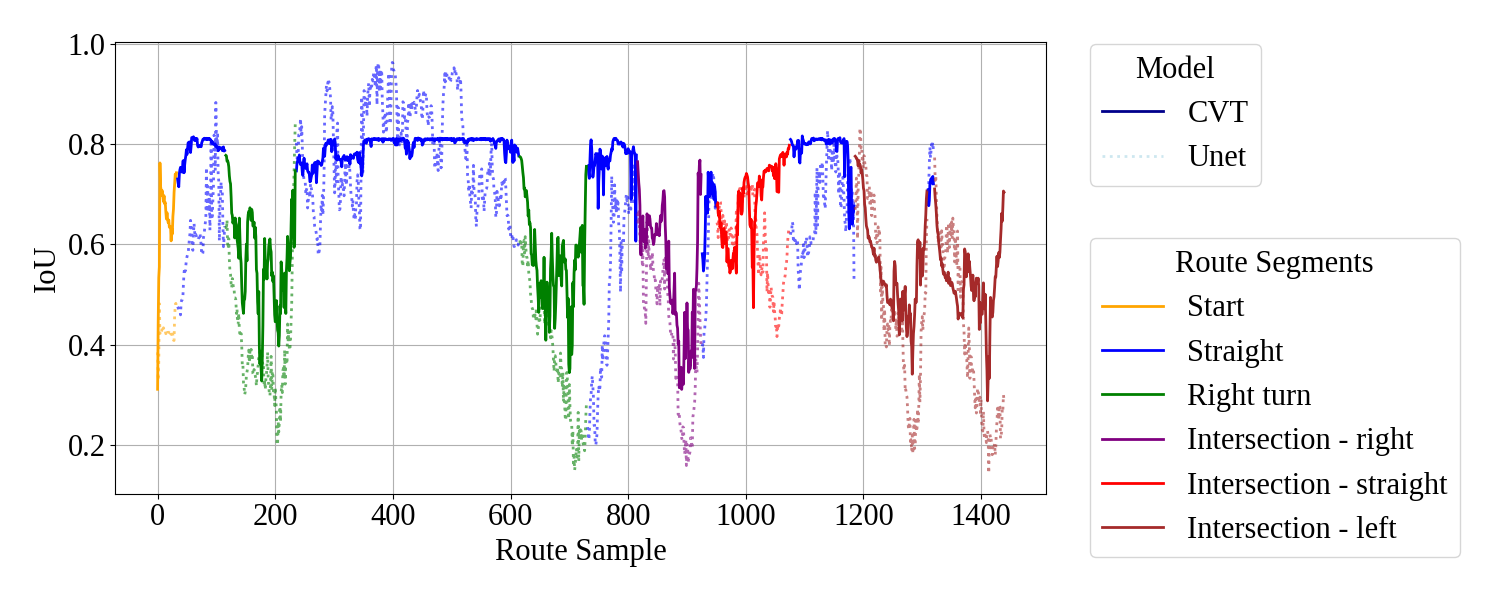} 
    \caption{
    \eric{IoU for CVT and Unet predictions} during part of the \textit{route00} from \textit{Town02}. For the colored segments of the Figure ~\ref{fig:test_envs}(b).}
    \label{fig:iou_route_cvt_unet} 
\end{figure}

As for the overall performance in the lane channel, the advantage still lies with the Unet models. One possible explanation is that these models perform better on straight-line segments. Considering that most of the routes are composed of straight trajectories, the mean IoU for Unet in this channel remains higher than that of the CVT models, which appear to struggle more with lane segmentation. This behavior is better illustrated in Figure~\ref{fig:iou_route_cvt_unet}.

Figure~\ref{fig:iou_route_cvt_unet} presents the mean IoU for each data point along the segments, using colors that match those in Figure~\ref{fig:test_envs}(b), enabling a comparison between CVT and UNet across different road sections. The transformer-based architecture demonstrates superior performance in turns and intersections. On the other hand, it appears to saturate around 0.8 on straight paths, while Unet occasionally surpasses this threshold. This suggests that Unet may struggle to infer from rarely seen situations but is more effective at reproducing familiar patterns. Otherwise, CVT shows a stronger ability to generalize and extract meaningful representations in more complex or uncommon scenarios, characteristic more aligned with the requirements of the BEV generation task.



\section{Conclusion}
\label{sec:conclusion}

This study investigates Bird’s-Eye View (BEV) generation \eric{for autonomous vehicles}
in the CARLA simulator using Cross-View Transformers (CVT). 
\eric{We included the trajectory channel to be generated, which is not covered in other works from the literature as far as the authors know. With this additional channel, an end-to-end imitation learning agent can be trained conditioned on the route to follow \cite{antonelo2024investigating}.}
Furthermore, the results demonstrate that the CVT architecture significantly outperforms the UNet baseline in terms of generalization, particularly in complex scenarios such as intersections and turns, especially for road and trajectory channels. A comparative analysis of loss functions revealed that CVT trained with L1 loss offers greater robustness in unseen environments, while UNet, though less adaptable, showed an advantage in lane segmentation and performance on straight road segments. 

The three-camera frontal configuration proved sufficient for most tasks in this work, reducing reliance on rear sensors without compromising performance. This suggests that autonomous systems in early deployment stages can benefit from data- and hardware-efficient solutions. However, lane segmentation remains a challenge for CVT models \eric{in the current setup}.

This work advances BEV perception research using the CARLA simulator by validating the performance of transformer-based models. 
\eric{However, further research should be done in training the models longer, with more data, and using resampling techniques, specially the ones with four cameras. For instance, data resampling could improve the performance of CVT models by weighting more the less frequent observations.}
Future work should also explore the direct integration of generated BEV representations into autonomous control pipelines, as well as the inclusion of more complex semantic channels such as pedestrians, \eric{other vehicles} and traffic lights.



%
%
%
\bibliographystyle{sbc}
\bibliography{biblio}

\end{document}